\documentclass[dvipsnames]{article}

\usepackage{arxiv}
\usepackage[numbers,compress]{natbib}

\usepackage[utf8]{inputenc}
\usepackage[T1]{fontenc}
\usepackage[hidelinks]{hyperref}
\usepackage{bookmark}
\usepackage{url}
\usepackage{booktabs}
\usepackage{amsfonts}
\usepackage{amsmath}
\usepackage{amssymb}
\usepackage{nicefrac}
\usepackage{microtype}
\usepackage{xcolor}
\usepackage{graphicx}
\usepackage{siunitx}
\usepackage{tikz}
\usetikzlibrary{arrows.meta,3d}
\usepackage{fontawesome5}

\makeatletter
\newcommand{\@crefprefix}[1]{%
  \edef\@crefarg{#1}%
  \expandafter\@crefdispatch\@crefarg::\@nil
}
\def\@crefdispatch#1:#2:#3\@nil{%
  \def\@crefkey{#1}%
  \ifx\@crefkey\@crefseckey Section\else
  \ifx\@crefkey\@crefappkey Section\else
  \ifx\@crefkey\@crefkeyfig Fig.\else
  \ifx\@crefkey\@crefkeytab Table\else
  \ifx\@crefkey\@crefkeyeq Eq.\else
  Section\fi\fi\fi\fi\fi
}
\def\@crefseckey{sec}
\def\@crefappkey{app}
\def\@crefkeyfig{fig}
\def\@crefkeytab{tab}
\def\@crefkeyeq{eq}
\providecommand{\cref}[1]{\@crefprefix{#1}~\ref{#1}}
\providecommand{\Cref}[1]{\@crefprefix{#1}~\ref{#1}}
\makeatother

\newcommand{\dz}{d_0}
\newcommand{\zz}{z_0}
\newcommand{\phii}{\varphi}
\newcommand{\thetaP}{\theta}
\newcommand{\qop}{q/p}

\newcommand{\kf}{KF}

\newcommand{\ttbar}{$t\bar{t}$}

\newcommand{\rms}{\ensuremath{\mathrm{RMS}}}
\newcommand{\cpuBaseline}{$\sim$170\,k}          %
\newcommand{\deployThroughputAligned}{5.41\,M}
\newcommand{\adaThroughput}{1.51\,M}

\usepackage{placeins}
\usepackage{algorithm}
\usepackage{algpseudocode}
\usepackage{float}

\title{Fast and Precise Learned Charged-Particle Trajectory Regression at the Large Hadron Collider}
\renewcommand{\shorttitle}{Fast and Precise Learned Charged-Particle Trajectory Regression at the LHC}
\date{}
\renewcommand{\undertitle}{}

\author{%
\begin{minipage}{\dimexpr\textwidth-24pt\relax}\centering
  \normalfont\normalsize
  {\bfseries
  Jonathan Renusch\textsuperscript{1,2},\ \
  Benjamin Huth\textsuperscript{1},\ \
  Daniel Murnane\textsuperscript{6,7}, 
  Eleni Xochelli\textsuperscript{1,3},\ \
  Do\u{g}a Elitez\textsuperscript{1,4},\\
  Paul Gessinger-Befurt\textsuperscript{1},\ \
  Andreas Stefl\textsuperscript{1},\ \
  Jeremy Couthures\textsuperscript{1,5},\ \
  Andreas Salzburger\textsuperscript{1},\\
  Lukas Heinrich\textsuperscript{2},\ \
  Michael Kagan\textsuperscript{8},\ \
  Markus Elsing\textsuperscript{1}}
  \\[6pt]
  \textsuperscript{1}CERN, Geneva, Switzerland \quad
  \textsuperscript{2}Technical University of Munich, Germany\\
  \textsuperscript{3}Universitat Aut\`onoma de Barcelona, Spain \quad
  \textsuperscript{4}Johannes Gutenberg-Universit\"at Mainz, Germany\\
  \textsuperscript{5}LAPP, Universit\'e Savoie Mont Blanc, CNRS/IN2P3, Annecy, France\\
  \textsuperscript{6} Niels Bohr Institute, University of Copenhagen, Denmark \quad 
  \textsuperscript{7} Lawrence Berkeley National Laboratory, USA \\
  \textsuperscript{8}SLAC National Accelerator Laboratory, USA\\
\end{minipage}
}

\begin{document}

\maketitle
\let\oldthefootnote\thefootnote
\renewcommand\thefootnote{}\footnotetext{ A Preprint. Correspondence to \texttt{jonathan.renusch@cern.ch}.}%
\let\thefootnote\oldthefootnote

\newcommand{\codepill}[1]{%
  \tikz[baseline=(B.base)]{%
    \node[draw=none, rounded corners=5pt,
          fill=black!14, inner xsep=10pt, inner ysep=5pt,
          font=\small] (B) {\faGithub\ \texttt{#1}};%
  }%
}
\begin{center}
\vspace{-25pt}
\href{https://github.com/jonathanrenusch/ssm-colliderml-track-regression}{%
  \codepill{Code}%
}
\vspace{15pt}
\end{center}

\begin{abstract}
We propose a training recipe
that treats charged-particle trajectory parameter regression on
high-energy physics detector data as a sequence-modeling task.
Kalman filters and linearized least-squares fits have been the classical standard
approach for this task: they are optimal estimators for sparsely sampled
linear-Gaussian data and are commonly used for trajectory parameter regression (fitting). The classical fitting techniques implemented for this domain reach a final precision of one part in $10^5$ through detailed modeling of detector geometry, material, detection effects and precise numerical integration of the equations of motion through the detector's inhomogeneous magnetic field.
With this study, we demonstrate that using a bidirectional gated linear recurrent encoder, one is able to reproduce the full precision of classical track fitting techniques. Using a custom kernel, we also achieve significantly higher throughput during GPU inference, compared to classical fitting software running on similarly priced multi-core CPU servers representing typically employed hardware. Such a speedup would lead to considerable cost savings for the pattern recognition at the Large Hadron Collider.
To our knowledge,
this is the first end-to-end learned track fit to reach the full precision and, at the same time, offer the opportunity to reduce the computing costs.
\end{abstract}

\section{Introduction}
\label{sec:intro}

\begin{figure}[t]
  \centering
  \includegraphics[width=\linewidth]{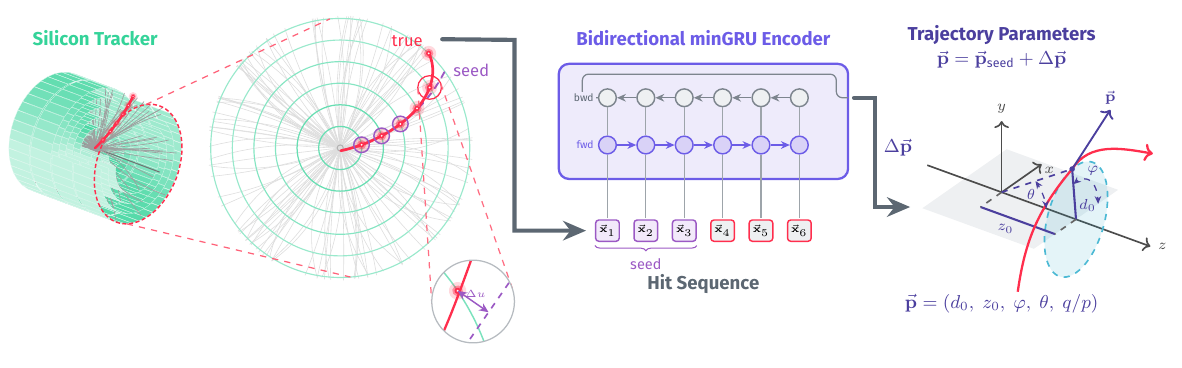}
  \caption[Charged-particle track fitting as sequence regression]{Charged-particle track fitting as sequence regression.
  A charged particle's hits across the silicon layers of the tracking detector~\citep{Gessinger2023ODD}
  form an ordered sequence of feature vectors $\vec{\mathbf x}_N$. An analytic helix through the first three hits gives a
  seed estimate $\vec{\mathbf p}_{\text{seed}}$; each hit carries measured coordinates plus three seed residuals:
  $\Delta u$, $\Delta v$, compressed with $\operatorname{asinh}$, and $s_{\mathrm{helix}}$. A bidirectional minGRU
  encoder reads the sequence and predicts a correction $\Delta\vec{\mathbf p}$ to the seed, giving the perigee parameters
  $\vec{\mathbf p}=(\dz,\zz,\phii,\thetaP,\qop)$ at closest approach to the collision axis. Schematic, not to scale.}
  \label{fig:workflow}
\end{figure}

Machine learning approaches to data processing problems in high-energy physics are an active field of R\&D.
Detecting the patterns of high-energy charged particles produced in proton--proton collisions at experiments at CERN such as ATLAS and CMS \cite{ATLASCollaboration2008Detector, CMSCollaboration2008Detector} and regressing their trajectory parameters is one of the most computationally complex tasks required to extract physics information from the massive datasets acquired  by the experiments (illustrated in \cref{fig:workflow}).

This charged particle ``tracking'' task is a ``connect-the-dots'' game: each particle traverses layers of detector material;
it scatters and deposits energy through ionization in particle detectors. Each such energy deposit in a detector we call a ``hit''. A pattern recognition algorithm has to recover which hits belong to which particle and what its trajectory was. The challenge grows in compute complexity when many particles arrive at once (\cref{fig:workflow}).   
At the High-Luminosity Large Hadron Collider (HL-LHC), which is scheduled to start taking data in 2030, every \SI{25}{\nano\second} a crossing of two proton bunches will give rise to on average 200 concurrent proton--proton interactions and in turn yield of the order of half a million hits across the detectors of the tracking sub-system~\citep{HLLHC2020TDR, ATLAS2017ITkPixel, CMS2017Tracker, Elitez2025ColliderML}. The pattern recognition algorithms reconstruct between 1000 and 2000 trajectories for such an event. Regressing the trajectory parameters with the highest precision for each particle trajectory is fundamental to achieving the physics goals of the experiments.

Computing cost is a central constraint on these plans. In 2022, the ATLAS experiment budgeted 8.8\,MCHF to extend the Event Filter for the HL-LHC upgrade of that compute hardware, assuming that CPU computing becomes about $15\,\%$ cheaper every year~\citep{ATLAS2022EFTracking}.
Since 2025, however, prices have moved the other way: driven by AI data-center demand, conventional DRAM contract prices roughly doubled in the first quarter of 2026 alone~\citep{TrendForce2026DRAM}, and the average selling price of servers without GPU accelerators rose by a third within a year~\citep{IDC2026Server}. These trends might become a serious long-term problem for the physics program of the LHC experiments, and they make algorithms that deliver more throughput per unit of hardware cost correspondingly more valuable.

The Kalman filter is a classical, optimal estimator for this regression task, which linearizes the problem using numerical field integration \citep{Fruhwirth2020Book} for the particle transport in the detector's magnetic field. A Kalman filter is
itself a state-space model~\citep{Fruhwirth1987KF, Strandlie2010RMP,
Fruhwirth2020Book}. This motivated our choice of encoder: its state update
has the same structure as the Kalman filter update, as detailed in
\cref{sec:encoder}. We use this correspondence as a guide for our design,
not as a claim that other encoders or further optimization could produce architectures achieving higher throughput. We tested four encoder families at matched parameter count under an identical
training recipe (\cref{fig:encoder-ablation}). Three of them, a bidirectional minGRU~\citep{Feng2024minGRU}, a
bidirectional Mamba-2~\citep{Dao2024Mamba2} and a  Transformer~\citep{Vaswani2017Attention}, reach the reference precision. A
parameter-matched diagonal state-space model without an input-dependent gate
fell short under the same protocol, most clearly in the momentum estimate.
Of the three, the bidirectional minGRU encoder gave
the highest throughput in our implementation (\cref{sec:kernels}), and we
use it for all main results. 

To our knowledge, this is the first learned trajectory fit to match classical Kalman-filter precision, offering the potential for significant cost savings through higher throughput at comparable hardware cost, and for replacing, within two days of training directly on simulated data, months of detector-specific software development usually required for a new detector design study.

\subsection{Related work}

Learned approaches to charged-particle
reconstruction at the LHC have so far mostly focused on the
``trajectory finding'' problem, the combinatorial assignment of
detector hits to candidate trajectories, rather than on
``trajectory fitting'', the regression of trajectory parameters at the
precision downstream physics requires.  The most-developed line uses
graph neural networks.  The ATLAS \textsc{GNN4ITk}
pipeline~\citep{ATLAS2024GNN4ITk} hands track candidates to the
classical Kalman filter for parameter estimation.
The complementary transformer line is exemplified by the
MaskFormer-style segmentation plus reconstruction architecture of
\citet{VanStroud2025Tracker}. It similarly targets joint hit
assignment and per-trajectory property estimation, but reports its
main metrics on assignment efficiency and fake rate.
The closest prior attempt at learned ``trajectory fitting'' on the
same detector is the transformer-based study of
\citet{Couthures2025CTD}, which targets parameter regression on the
Open Data Detector~\citep{Gessinger2023ODD} with full ACTS~\citep{Ai2022ACTS} simulation; the reported
per-parameter resolutions approach the Kalman filter results without matching
them.

Kalman filter fitting also runs on GPUs in production where the
propagation step admits a detector-specific shortcut, e.g.\ at LHCb
and ALICE experiments~\citep{billoir2026fullrate, rohr2019alicetpc}. For
general-purpose silicon trackers, where material effects and field
integration cannot be simplified this way, GPU-based Kalman fitting remains
a topic of R\&D with only modest gains in computing costs so far~\citep{yeo2025traccc}. The
obstacle is structural: the propagation between hits is data-dependent
(which surface comes next, how far to step, which material to
cross) and is, by itself, an iterative numerical field integration task.
It does not reduce to one dense kernel shared by every
trajectory, and GPU threads diverge instead. Our learned
fit has no propagation step at inference: geometry, material and
field are absorbed into the weights during training, so every trajectory
regression results in the same dense computation task and parallelizes trivially, at
Kalman filter precision.

\phantomsection\label{sec:related}

\section{Data}
\label{sec:data}

We use the \textsc{ColliderML}
dataset~\citep{Elitez2025ColliderML, ColliderMLHF}, publicly available at
\href{https://huggingface.co/datasets/CERN}{huggingface.co/datasets/CERN}, an open-source dataset
of fully simulated $\sqrt{s}=14$~TeV proton--proton
physics (collisions with a $14$~TeV center-of-mass energy
between the two colliding proton bunches) on the ``Open Data Detector''
(ODD)~\citep{Gessinger2023ODD}, an open community
silicon-tracker geometry to facilitate algorithmic R\&D (cutaway in
\cref{fig:workflow}).  The detector sits inside a solenoid magnet that produces an almost uniform \SI{3}{\tesla} magnetic field along the collision axis, which forces charged particles onto
curved trajectories and in turn allows their momentum to be inferred from that
curvature. As this work focuses only on the regression problem within the
track reconstruction chain, we assume perfect hit-to-track matching, so the
learned minGRU and the baseline Kalman filter both receive exactly the
same sequence of hits. 

Each silicon hit contributes $12$ measured features (position,
derived angles, and detector identifiers); three further per-hit features derived from an
analytic helix seed are described in \cref{sec:method}. The
positions measured at the sensor were overlaid with Gaussian smearing, in
order to simulate the actual detection precision. Hits are
ordered by the time at which they were measured, which gives the learned minGRU a meaningful inductive
bias about positional information, letting it retrace the original
trajectory the particle took through the detector.

The training set comprises $387$\,M tracks: $372$\,M from single-muon samples, uniform in
transverse momentum $p_{\mathrm T}\in[1,110]$~GeV ($191.5$\,M) and
log-uniform in $p_{\mathrm T}\in[0.9,110]$~GeV
($180.0$\,M, supplying low-momentum statistics), and $15.8$\,M ($4\,\%$) from \ttbar{} events, which are statistically independent of the \ttbar{} test sample. The tracks are required to
have $|\eta|\le 3$ and $6$--$20$ hits in the tracking system\footnote{$\eta = -\ln{\tan{\frac{\theta}{2}}}$ is the
``pseudorapidity'' calculated from polar angle $\theta$, a standard collider-physics feature that measures
the particle's direction relative to the collision axis ($\eta=0$
is perpendicular to the beam, larger $|\eta|$ means closer to the
collision direction).}. Muons are elementary particles that serve as the field's standard calibration candle for any new charged-particle reconstruction algorithm.

Evaluation uses four held-out muon samples disjoint from all training
data: single muons at fixed $p_{\mathrm T}$ of $2$, $10$, and
$50$~GeV ($10^5$ tracks each) and a uniform $1$--$70$~GeV muon
spectrum ($3$\,M tracks), together with a held-out \ttbar{} sample ($p_{\mathrm T}>1$~GeV) that adds charged hadrons and electrons. The fixed-$p_{\mathrm T}$ samples cover the range from the low
momentum regime ($2$~GeV), in which multiple scattering (deflection of the particle by the detector material) is the dominating effect, to the high momentum regime ($50$~GeV) for which the precision is only limited by the intrinsic resolution of the detector measurements.

The standard metric for evaluating a
track fitter in particle physics is an iteratively
$3\sigma$-clipped \rms{} of the residuals, the spread of the
central body of the residual distribution after a fixed-rule
outlier rejection. It is calculated by iteratively clipping away residuals lying outside of a $3\sigma$ range until the distribution stabilizes. This is not the plain \rms{} the wider
machine-learning literature usually reports; on heavy-tailed data
the plain \rms{} is dominated by a small number of outliers
and obscures the behavior of the model on the bulk of the sample. Resolution-curve uncertainties are the analytic \rms{} standard error per bin.

\section{Method: A seed-guided bidirectional minGRU trajectory parameter estimator}
\label{sec:method}

\begin{figure}[t]
  \centering
  \includegraphics[width=0.50\linewidth]{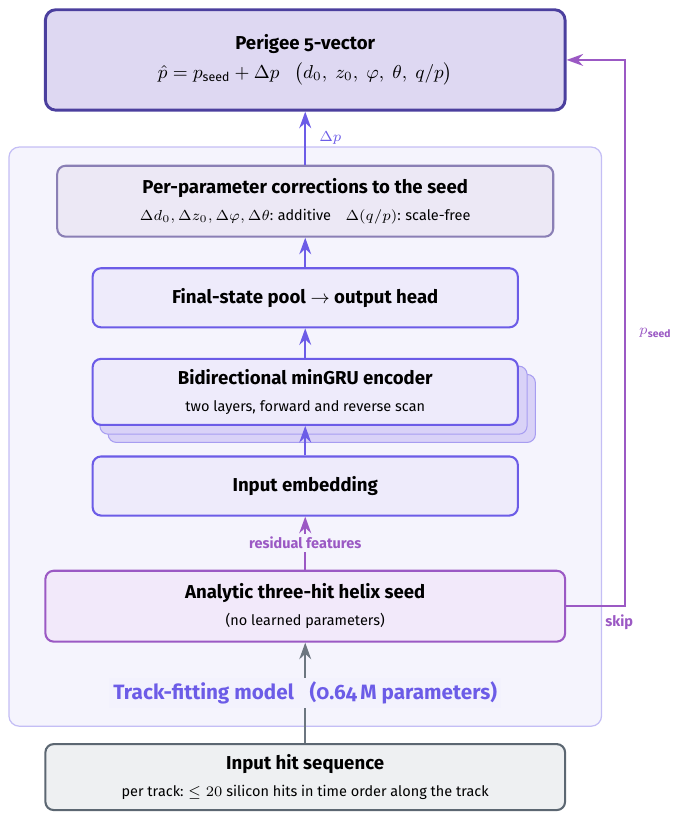}
  \caption{Architecture of the seed-guided bidirectional minGRU model.  An analytic three-hit
  helix seed (top, no learned parameters) supplies residual features
  to the encoder input and, via the skip connection, anchors the
  output.  The encoder ($0.64$\,M parameters) predicts per-parameter
  quantile corrections $\Delta p$ (7 quantiles each, median as the
  point estimate); the final estimate is
  $\hat p = p_{\mathrm{seed}} + \Delta p$, additive for
  $\dz,\zz,\phii,\thetaP$ and a scale-free residual for $\qop$.}
  \label{fig:architecture}
\end{figure}

The model maps the input hit sequence of a single track, ordered by hit measurement time, to the five perigee
parameters of that track, $(\dz, \zz, \phii, \thetaP, \qop)$ that describe the trajectory at its closest approach to the nominal collision axis (\cref{fig:workflow}). The regression target is defined as the deviation from an initial ``seed'' perigee estimate, which is analytically derived from the first three hits of the trajectory. The seed perigee together with the deviations per hit allows the model to predict a correction to the seed trajectory, in a similar fashion to how a Taylor-expanded Kalman filter implementation fits for the linear correction term to the ``seed'' trajectory \citep{Fruhwirth2020Book}. \cref{fig:architecture} summarizes the architecture of the seed-guided bidirectional minGRU model. 

\subsection{Analytic seed}

A seeding trajectory estimate constructed from at least three hits is commonly used in classical trajectory finding and reconstruction techniques~\citep{Ai2022ACTS}.  We use the seed trajectory for this study because anchoring the
regression targets to the seed collapses their wide dynamic range into a
residual of comparable scale at every momentum, keeping the model's
internal representations well conditioned. Consequently, the same precision of the regression results can be achieved using far fewer parameters, in turn resulting in faster
inference, compared to a model that learns the absolute value directly. Concretely, a model that learns the absolute value of the perigee parameters without a seed requires roughly $8\times$ more parameters to achieve the same precision as the model incorporating the seed.

A trajectory of a charged particle in an (approximately) homogeneous \SI{3}{\tesla} magnetic field of the Open Data Detector is described by a helix only in the ideal case. The three hits of a trajectory seed are sufficient to analytically calculate all five seed perigee parameters using a conformal (inversion) map~\citep{Karimaki1991Circle} that turns
``find the circle through three points'' into ``draw the straight line
through two points'' via elementwise closed-form arithmetic, with no fit or iteration involved.\footnote{The seed is a vectorized port of the ACTS track-parameter estimator~\citep{Ai2022ACTS} (\texttt{estimateTrackParamsFromSeed}, the conformal-map estimate of the ATLAS experiment's~\citep{ATLASCollaboration2008Detector} silicon seed maker), combined with the triplet rule of the ACTS truth-seeding algorithm. It is closed-form and computed independently per track from $(x,y,z)$, the detector-volume identifier and the constant $B_z = \SI{3}{\tesla}$, nothing else.} The circle's curvature and turning direction give
$p_{\mathrm T}$ and the charge, and the direction w.r.t. the collision axis extracted from the same hits gives $\cot\thetaP$. Extrapolating the result geometrically to the collision axis defines the perigee (closest point of approach in the plane perpendicular to the collision axis) at which all five parameters
$(\dz,\zz,\phii,\thetaP,\qop)$ are given. The only inputs to this procedure are the measured hit positions and the
detector-volume identifiers: no learned
parameters, no geometry or material description. The identical
closed-form computation runs inside
the model's own forward pass on the GPU. Technically the seed perigee is computed in float64 rather than the
surrounding float32 because of the need to avoid a cancellation error in the
beamline transport that becomes significant at high momentum.

The seed perigee then enters the
model twice. On the input side every hit gains three features: its
two signed distances from the seed helix, decomposed in the curvilinear
frame the conventional linearized Kalman filter~\citep{Fruhwirth2020Book} operates in as well and compressed
with $\mathrm{asinh}$ to tame their tails ($\mathrm{asinh}\,\delta u$, $\mathrm{asinh}\,\delta v$), plus its path length along the helix, $s_{\mathrm{helix}}$. On the output
side every head predicts a correction to the seed rather than an
absolute value, a target-side skip connection through which no
gradient flows.

\subsection{Encoder}
\label{sec:encoder}

The $15$ per-hit features are min--max
normalized to $[0,1]$ and expanded by a multi-scale Fourier
featurization with $16$ scales~\citep{Tancik2020Fourier}, giving a $480$-d per-hit vector projected through a
dense network to the encoder width $d_{\mathrm{model}} = 128$.  The
backbone stacks two bidirectional minGRU layers~\citep{Feng2024minGRU} with
a hidden width of $192$ ($0.64$\,M parameters in total). Each layer
linearly projects its input to the forward and reverse gates and candidate
states, then runs a forward and a reverse scan. The two directions' states
are concatenated ($384$-d) into the next layer.

The Kalman filter is efficient because it processes the hits one at a time with a small state, so its cost grows only linearly with the number of hits~\citep{Fruhwirth2020Book, Fruhwirth1987KF}. The minGRU shares this structure:
its recurrence mirrors the state-update logic of the Kalman
filter applied along the trajectory. At every hit, the Kalman filter
first propagates a predicted state $\vec x_{\,t-1}$ from the
previous estimate, then corrects it with the gain $\mathbf K_t$
applied to the difference between the measurement $\vec m_t$ and its
projection $\mathbf H_t \vec x_{\,t-1}$ into measurement space. Its core assumption is that both process noise (material effects) and measurement noise can be described as Gaussians. The
per-hit minGRU update propagates a hidden state $\vec h_{t-1}$
corrected by an input-dependent gate $\vec z_t$ applied to the
difference between a candidate state $\vec n_t$, computed from the embedded
measurement $\vec m_t$, and the previous state. The state updates in a Kalman filter and the minGRU model are
\begin{align}
\vec x_t &= \vec x_{\,t-1} \;+\; \mathbf K_t\bigl(\vec m_t - \mathbf H_t\,\vec x_{\,t-1}\bigr),
   \label{eq:kalman-update} \\
\vec h_t &= \vec h_{t-1} \;+\; \vec z_t \odot \bigl(\vec n_t - \vec h_{t-1}\bigr).
   \label{eq:mingru-update}
\end{align}

Here $\odot$ denotes an elementwise (Hadamard) product, following the
minGRU formulation~\citep{Feng2024minGRU}. Concretely, $\vec z_t = \sigma(\mathrm{Linear}(\vec m_t))$ is a per-channel,
input-dependent gate and $\vec n_t = \mathrm{Linear}(\vec m_t)$ a candidate
state. This gate plays the role of the Kalman gain
$\mathbf K_t$ in the classical Kalman filter track fit, and, more loosely,
of the explicit field integration and material-effect corrections that feed
$\mathbf K_t$ there, but it is learned from data rather than derived from a
complex model.
The correspondence is not exact. The Kalman gain is computed from the
uncertainty of the current estimate, and therefore depends on all previous
hits along the trajectory. The minGRU gate depends only on the current hit,
which is what allows all gates to be precomputed.

One default minGRU layer scan is forward-only while the Kalman filter runs a forward as well as reverse scan~\citep{Ai2022ACTS}. This is why we make the minGRU bidirectional per layer: so the hidden state at each hit is informed by both its past and its future along
the trajectory.

The final state of the forward scan (at the outermost hit) and the final
state of the reverse scan (at the innermost hit) are concatenated to a
$384$-d representation, normalized~\citep{Zhang2019RMSNorm} and read by a
two-layer head (hidden width $128$) producing $7$ quantiles for each of the
five parameters, $35$ outputs in total.

\subsection{Losses}

For each of the five parameters the model predicts seven quantiles of the residual distribution, trained with the standard quantile loss~\citep{Koenker1978Quantile} at equal weight, and the median is taken as the estimate. The losses are applied
to the seed-anchored targets: $\Delta\dz$ and $\Delta\zz$ within
$\pm 0.4$ and $\pm\SI{3.5}{\milli\metre}$, $\Delta\phii$ (wrapped)
and $\Delta\thetaP$ within $\pm 15$ and $\pm\SI{10}{\milli\radian}$, all four simple additive
offsets to the seed, and for $\qop$ a ``scale-free'' residual
$(\qop - \qop_{\mathrm{seed}})\,/\,(|\qop_{\mathrm{seed}}| +
\epsilon)$ with $\epsilon = 0.02$~GeV$^{-1}$.  The scale-free form
makes a $1$~GeV and a $100$~GeV trajectory place the same relative demand
on output precision. If the model predicted $\qop$ directly, the residual would be only
$\sim\!10^{-4}$ of the output range, too small to be learned reliably.

\subsection{Training}

Training runs end-to-end in strict fp32, giving comfortable headroom in the mantissa to
the $10^{-5}$ precision this problem demands. 
The batch size is deliberately small at $2048$: a $36\,000$ batch loses up to a factor of two on the
azimuthal and curvature resolutions. Small batches are known to give more precise results on unseen data, an effect attributed to the extra noise in each update step~\citep{Keskar2017LargeBatch,Smith2018BayesianSGD,Smith2020NoiseBenefit}, and the same has been reported for other high-precision regression tasks~\citep{Batzner2022NequIP,Borovykh2019Forecasting}. The
model is trained in two phases. The first runs $25$ passes over the $387$\,M-trajectory mixture with the Lion optimizer~\citep{Chen2023Lion} ($\sim\!31$~h on one H100~\citep{NvidiaH100Datasheet}). The second runs another $25$ passes with $2\times20\,000$ tracks per step on two H100s ($\sim\!15$~h), using the Muon and AdamW optimizers~\citep{Jordan2024Muon} while the step size is brought down to zero~\citep{Hu2024MiniCPM}. This second phase leaves the clipped core resolutions unchanged, reduces the residual tails, and closes a final resolution gap of $\approx 3\,\%$ on $\qop$.

\subsection{Kernel adaptation}
\label{sec:kernels}

Standard GPU kernels for sequence models target sequences of
$10^{3}$ to $10^{5}$ elements. A charged-particle trajectory has at most
$20$ hits and $13$ on average. A kernel built for the longer regime
wastes computation and start-up time on a track.

We give every encoder we tested the same treatment. Tracks run in a
packed, unpadded layout. For every encoder
the step that combines information across the hits of a track runs as a single GPU program (a fused Triton kernel). For the recurrent encoders that kernel scans
along the track and updates each channel independently. For the
Transformer the bias, activation, normalization and residual
addition around the Transformer's matrix multiplications are folded
into those multiplications. The per-hit Fourier encoding ahead of the encoder is compiled into a few
fused kernels. The dense matrix multiplications of the encoder layers, which map each hit's feature vector to the next layer, run in
fp16 for all three encoders. The recurrence itself
accumulates in fp32 inside the kernel (\cref{app:kernels}), and the seed
stays float64.
This treatment improves throughput by
$6.1\times$ for the minGRU, $6.6\times$ for the Transformer, and
$3.6\times$ for Mamba-2 (in tracks/s; \cref{tab:kernel-gains}, \cref{app:kernels}), measured on an H100 GPU. For
the non-selective state-space model we trained, no kernel optimization is
attempted, since it fell short of the reference precision under our training recipe
(\cref{fig:encoder-ablation}).

All encoders are compared at matched parameter count ($\sim$$0.63$\,M), with
an identical recipe and a shared learning rate. We did not tune depth, width
or learning rate per encoder, so the throughput ordering reflects our
implementations and may change under further per-architecture optimization.

\section{Results}
\label{sec:results}

This work advances the field along three main directions.
First, the minGRU
matches the mathematically optimal classical baseline~\citep{Fruhwirth1987KF, Strandlie2010RMP, Fruhwirth2020Book} (a Kalman filter) measured within an established validation framework~\citep{Ai2022ACTS} in absolute precision. This demonstrates that the model is able to learn the full complexity of the detailed modeling of classical trajectory fitting software in terms of detector geometry, material effect corrections and magnetic-field integration to achieve the $10^{-5}$ relative precision the problem demands from data. Second, the fit is embarrassingly parallel across tracks: with a GPU kernel adapted to the special structure of tracking data, our fastest model fits \adaThroughput{} tracks per second on an NVIDIA RTX 5000 Ada GPU. This demonstrates that the minGRU trajectory parameter estimator is able to achieve a significantly higher throughput compared to a classical Kalman filter trajectory fit run on a Threadripper 3970X 32-core CPU machine, which costs about half as much and fits roughly \cpuBaseline{} tracks per second. Third, the model learns a detector from its simulated data alone in about two days of training on two H100 GPUs: no hand-coded geometry description, material map, or field integration enters the parameter estimation model. For detector design studies for the Future Circular Collider~\citep{Abada2019FCC}, the training of a seed-guided bidirectional minGRU trajectory parameter estimator may replace the complex implementation work the classical trajectory-fitting approach requires for every candidate detector geometry, allowing for a faster turnaround in detector performance studies.

The result rests on three ingredients: training with small batches in 32-bit floating point, which improves the final precision; predicting a correction to an analytic seed track, together with quantiles of its uncertainty; and a custom GPU kernel adapted to the short hit sequences of tracks.

\subsection{Physics performance in precision}
\label{sec:results-headline}

\cref{tab:ratios} reports the ratio of the minGRU's resolution to the
truth-seeded \kf{}'s for all five perigee parameters on the four muon
test samples, for the clipped core and
for the un-clipped \rms{} that includes every tail.  The minGRU is at or
below the reference in every entry of both halves of the table within \SI{1.5}{\percent}.  The un-clipped results include
non-Gaussian tails in the data which the classical linear-Gaussian estimator is not handling fully appropriately.

\begin{table}[t]
  \centering
  \caption{Resolution ratio minGRU\,/\,truth-seeded \kf{} per perigee
  parameter and test sample ($|\eta|\le2$): iteratively
  $3\sigma$-clipped \rms{} (left) and un-clipped \rms{} including all
  tails (right).  Values $<1$ mean the minGRU achieves better resolutions than the
  reference. The reference is the truth-seeded \kf{} shipped with the
  dataset in every row.  Brackets are $1\sigma$ bootstrap uncertainties on the
  last digit, from $400$ paired replicas per sample.}
  \label{tab:ratios}
  \small
  \setlength{\tabcolsep}{3.4pt}
  \begin{tabular}{l ccccc c ccccc}
    \toprule
    & \multicolumn{5}{c}{clipped core (iter-$3\sigma$)} &
    & \multicolumn{5}{c}{un-clipped (with tails)} \\
    \cmidrule{2-6}\cmidrule{8-12}
    sample & $\dz$ & $\zz$ & $\phii$ & $\thetaP$ & $\qop$ &
           & $\dz$ & $\zz$ & $\phii$ & $\thetaP$ & $\qop$ \\
    \midrule
    $\mu$, 2\,GeV        & 0.992(2) & 0.991(3) & 0.989(2) & 0.993(2) & 1.006(4) & & 0.989(2) & 0.981(4) & 0.993(3) & 0.984(6) & 0.959(4) \\
    $\mu$, 10\,GeV       & 0.993(1) & 0.995(2) & 0.992(2) & 0.991(2) & 1.003(3) & & 0.968(5) & 0.950(5) & 0.969(5) & 0.986(2) & 0.987(3) \\
    $\mu$, 50\,GeV       & 0.998(1) & 0.999(1) & 0.977(2) & 0.994(3) & 1.013(2) & & 0.993(1) & 0.957(4) & 0.977(1) & 0.9968(8) & 1.018(4) \\
    $\mu$, 1--70\,GeV    & 0.9955(3) & 0.9981(3) & 0.9815(6) & 0.9958(5) & 1.0154(7) & & 0.980(3) & 0.960(5) & 0.975(7) & 0.98(1) & 0.96(2) \\
    \bottomrule
  \end{tabular}
\end{table}

\begin{figure}[!htbp]
  \centering
  \includegraphics[width=0.86\linewidth]{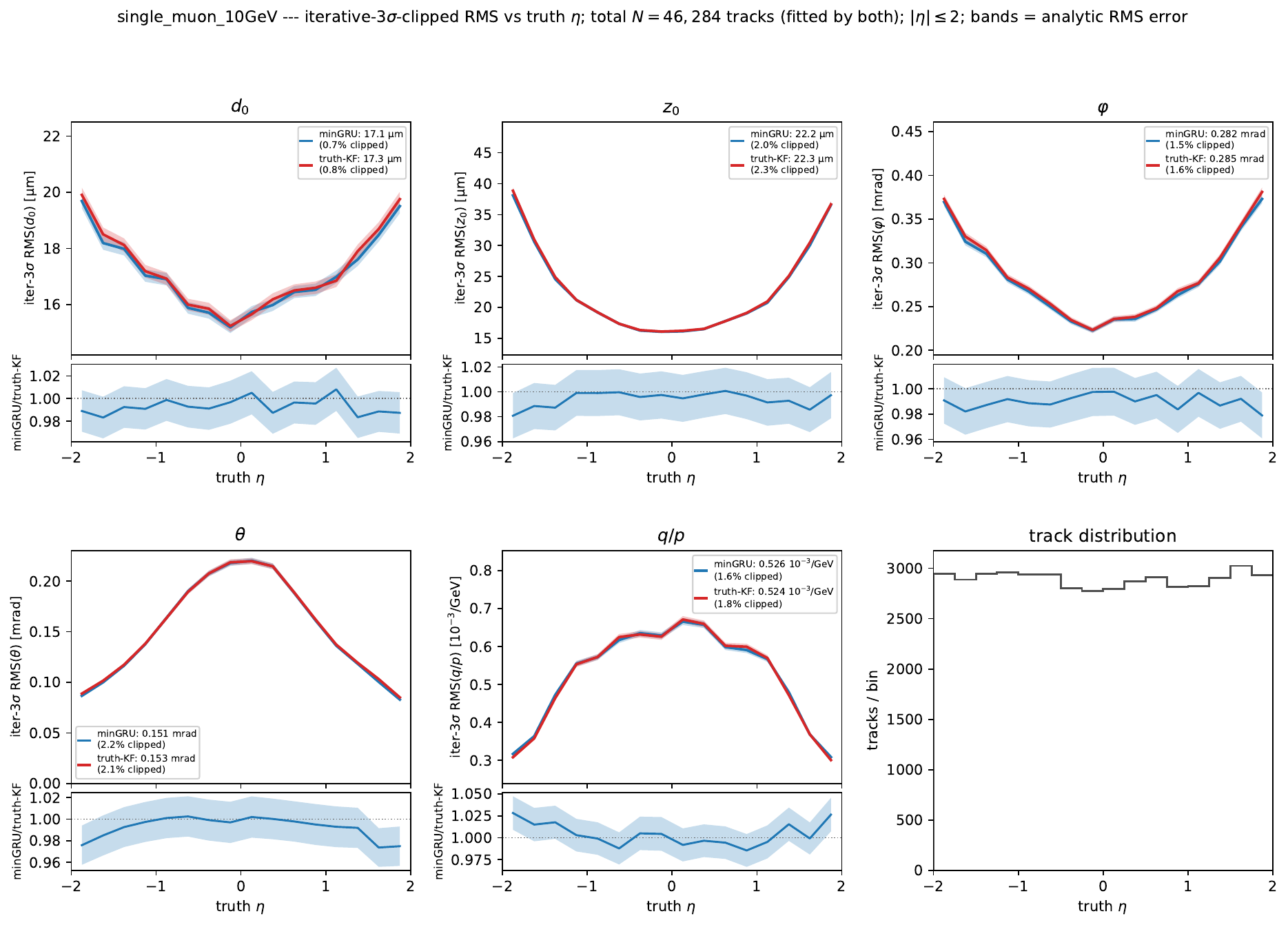}
  \caption{Core resolution versus pseudorapidity on the \SI{10}{\GeV}
  muon sample ($|\eta|\le2$): iterative-$3\sigma$-clipped \rms{} of
  each perigee parameter for the minGRU (blue) and the truth
  \kf{} (red), with the minGRU/\kf{} ratio
  beneath each panel.  Each legend gives the unbinned \rms{}; the title states the total
  number of trajectories fitted by both estimators.  Bands are the analytic \rms{}
  standard error.
  }
  \label{fig:rms_eta}
\end{figure}

\begin{figure}[!htbp]
  \centering
  \includegraphics[width=0.86\linewidth]{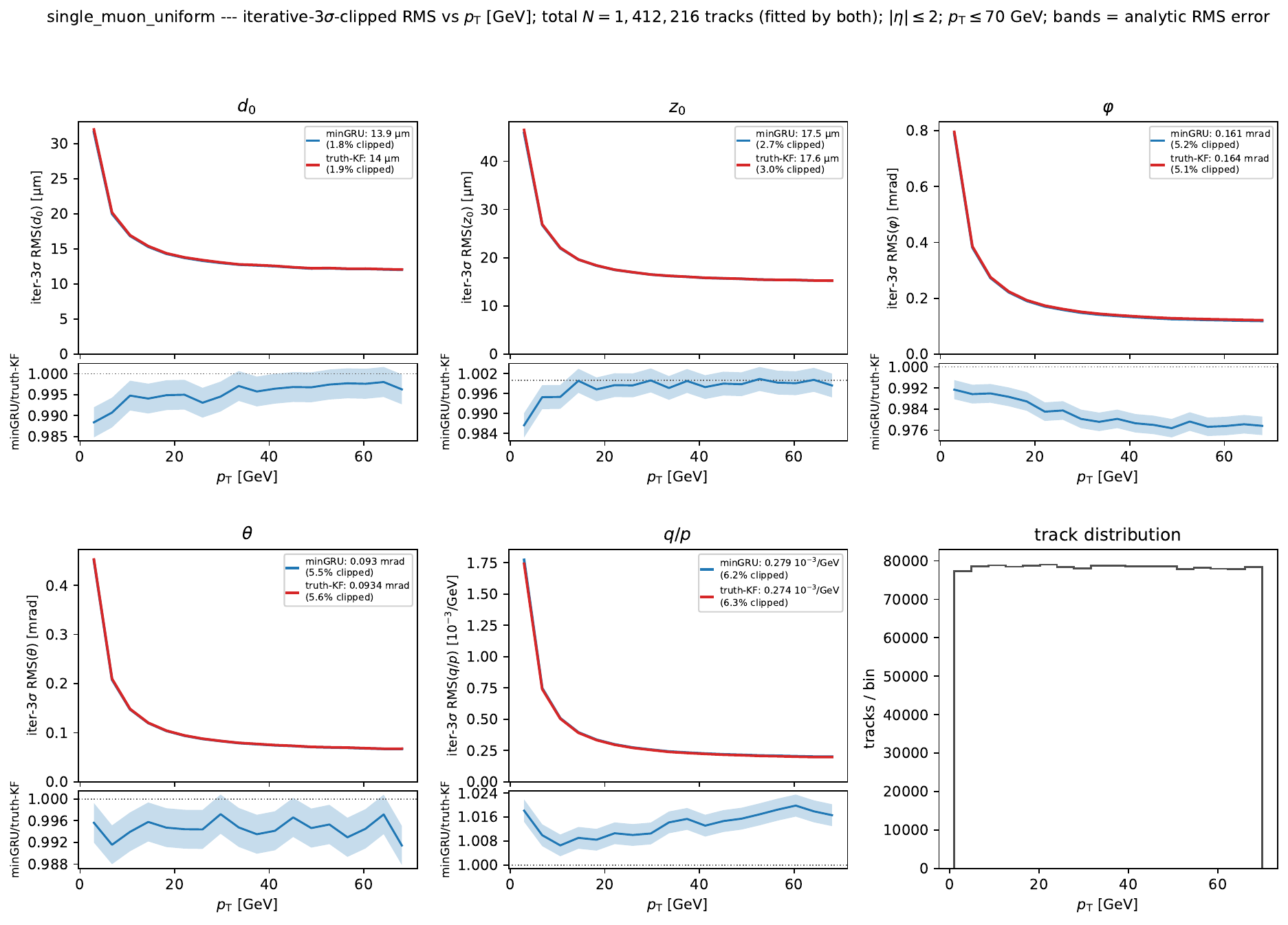}
  \caption{Core resolution versus transverse momentum on the uniform
  $1$--$70$\,GeV muon sample ($|\eta|\le2$, equal-width
  $p_{\mathrm T}$ bins): iterative-$3\sigma$-clipped \rms{} of each
  perigee parameter for the minGRU (blue) and the truth \kf{} (red),
  with the minGRU/\kf{} ratio beneath each panel.  Each legend gives the
  unbinned \rms{}; the title states the total number of trajectories fitted by both
  estimators.  Bands are the analytic \rms{} standard error.  %
  }
  \label{fig:rms_pt}
\end{figure}

\cref{fig:rms_eta} shows the core resolution versus pseudorapidity on
the \SI{10}{\GeV} muon sample: the two curves track each other, the minGRU at or marginally below the \kf{} in
every parameter across the full range studied. Integrated over the sample, the minGRU
reaches \SI{17.1}{\micro\metre} on $\dz$, \SI{22.2}{\micro\metre} on
$\zz$, \SI{0.28}{\milli\radian} on $\phii$, \SI{0.15}{\milli\radian}
on $\thetaP$ and $5.2\times10^{-4}$~GeV$^{-1}$ on $\qop$ over the
$46\,284$ tracks estimated by both, indistinguishable from or
marginally below the classical fit.  At \SI{50}{\GeV} and \SI{2}{\GeV} we see similar results, and
the resolution holds evenly across the full momentum range as well:
\cref{fig:rms_pt} shows the same clipped core resolution versus
transverse momentum on the uniform $1$--$70$\,GeV muon sample, where
the minGRU again tracks the truth-seeded \kf{} in every $p_{\mathrm T}$
bin of every parameter.  The \SI{2}{\GeV} and \SI{50}{\GeV}
resolution-versus-$\eta$ curves and the residual distributions are in
\cref{app:results-supplement}.
An independently held-out \ttbar{} sample shows similar performance (\cref{fig:app-ttbar}).

The trained minGRU encoder also recovers a covariance-like structure of
perigee parameter estimates without any supervision on it, consistent with the
physical correlations we would expect: the per-parameter loss
gradients on the shared trunk align within the geometric parameters
$(\dz,\zz,\phii,\thetaP)$ and are nearly orthogonal to $\qop$; see
\cref{app:interp-grad}.

\subsection{Throughput}
\label{sec:results-throughput}
\phantomsection\label{sec:results-interp}%

\begin{table}[t]
  \centering
  \caption{The throughput and throughput per device dollar.
  GPU rows: this work's model on the deployment path, at the saturating batch size
  (\cref{fig:throughput}).  Prices are the device-only list prices
  (GPU board / CPU chip \citep{ChipCost}; hosts excluded on every row). The gain of the kernel
  treatment for each encoder is given in \cref{tab:kernel-gains}.}
  \label{tab:throughput}
  \small
  \begin{tabular}{llrrr}
    \toprule
    device & configuration & tracks / s & approximate cost [\$] & ratio \\
    \midrule
    32-core CPU  & ACTS Kalman filter fit (reference)  & \cpuBaseline & 2\,000 & 86 \\
    RTX~5000~Ada & this work, deployment path  & \adaThroughput & 4\,000 & 378 \\
    H100~NVL     & this work, deployment path  & \deployThroughputAligned & 30\,000 & 180 \\
    \bottomrule
  \end{tabular}
\end{table}  

Each of the three encoders we gave a fused kernel improves throughput by between $3.6\times$
and $6.6\times$ over its default kernel, and in throughput the minGRU leads the Transformer
by $1.7\times$ and Mamba-2 by $2.6\times$ (\cref{tab:kernel-gains}), all numbers measured on an H100 NVL GPU.

\cref{tab:throughput}
gives a comparison of the throughput of the learned trajectory parameter estimator on a workstation RTX~5000~Ada GPU, a datacenter H100 GPU, and a Threadripper 3970X 32-core CPU. The H100 outperforms the workstation GPU in throughput. The model achieves a significantly higher %
throughput than running the classical ACTS Kalman filter trajectory fitting using all cores of a Threadripper 3970X CPU, even if normalized to the approximate cost of the compute device. The enormous data volumes produced by high-energy physics experiments require the collaborations to use the most cost-effective compute for processing the data. Taking into account the approximate price of the GPUs, the H100 loses its advantage to the RTX~5000~Ada GPU. %

The analytic seed is part of the deployed forward pass, so we time
it explicitly. At the saturating batch size, the full minGRU forward
costs \SI{0.19}{\micro\second} per track on the H100. The on-GPU
seed, computed in float64, takes \SI{0.016}{\micro\second} per track,
$9\,\%$ of the forward pass on the H100. For the RTX~5000~Ada, the full forward pass takes \SI{0.72}{\micro\second} and the on-GPU seed takes \SI{0.089}{\micro\second}  per track, $12\,\%$. The Fourier encoding and the quantile heads take about $50\,\%$, and the encoder itself $40\,\%$ of the full forward pass.  

\begin{figure}[!htbp]
  \centering
  \includegraphics[width=0.62\linewidth]{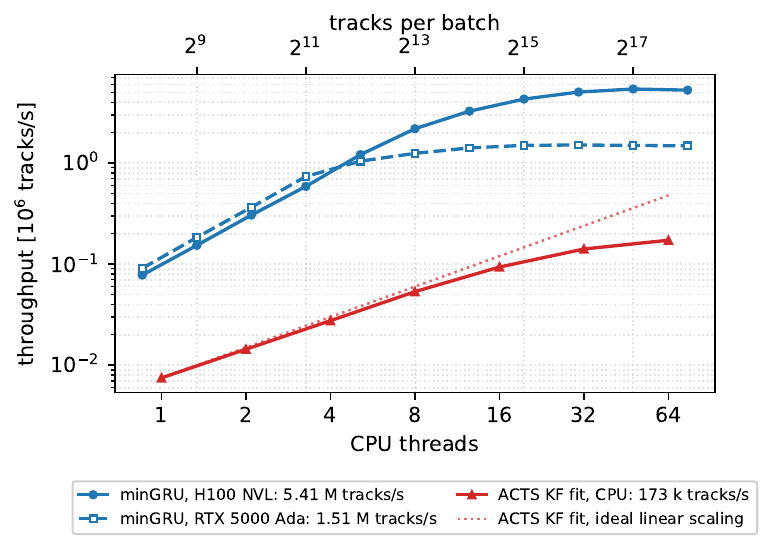}
  \caption{Inference throughput versus batch size (blue, top axis) on a
  datacenter H100~NVL, for the seed-guided bidirectional minGRU trajectory
  parameter estimator at fp16 inference.  The on-GPU seed calculation is
  allowed for in every throughput estimate.  The red curve is the ACTS \kf{}
  fit on an AMD Threadripper~3970X versus the number of threads it was given
  (bottom axis), reaching \cpuBaseline{} tracks per second at 64 threads; the
  dotted line is ideal linear scaling from the single-thread measurement.
  The dashed blue curve is the same model and benchmark on a workstation
  RTX~5000~Ada.}
  \label{fig:throughput}
\end{figure}

The batch-size dependence underlying the numbers below, on both a
datacenter H100~NVL and a workstation RTX~5000~Ada, is shown in
\cref{fig:throughput} (\cref{app:kernels}). Throughput saturates from
$\sim$$16$--$64$\,k tracks per batch on both GPUs. For comparison, the throughput on the Threadripper 3970X 32-core CPU is also shown as a function of the number of threads.

\section{Conclusions}
\label{sec:conclusion}\phantomsection\label{sec:future}%

We demonstrate that a seed-guided bidirectional minGRU,
a gated linear recurrent model, matches the precision of a mathematically optimal estimator, the classical Kalman filter trajectory fitter, for both the Gaussian core and the un-clipped tail distributions. Furthermore, a minGRU run on a workstation GPU yields significantly higher throughput compared to a classical Kalman filter implementation run on a cost-comparable multi-core CPU. The seed-guided bidirectional minGRU trajectory parameter estimator is capable of learning the detector geometry, material effect corrections and the charged-particle transport in the magnetic field from the simulated data itself and does not require complex hand-coded implementations, which for
detector-design studies such as a Future Circular Collider replaces months of software engineering and may therefore accelerate detector design studies.

It seems plausible that variants of the seed-guided bidirectional minGRU trajectory parameter estimator presented in this paper may be optimized for electron trajectory fitting, which is complicated by the effects of Bremsstrahlung in the detector material. Learning the joint distribution of non-Gaussian physical effects, which are particularly prominent for electrons, might enable significant precision gains over classical methods. Classical combinatorial Kalman-filter-based trajectory finding strategies use mathematical models similar to the minGRU version presented here. Further adapting the minGRU model may allow it to emulate the combined classical track finding and fitting chain.

\section*{Acknowledgments}
J.R., B.H., and P.G.B. are supported by the Eric \& Wendy Schmidt Fund for
Strategic Innovation through the CERN Next Generation Triggers project under
grant agreement number SIF-2023-004. L.H. is supported by BMFTR Project SciFM 05D25WO2.
D.M. was supported in this work by the Danish Data Science Academy, which is funded by the Novo Nordisk Foundation (NNF21SA0069429).

\paragraph{Conflict of interest.}
The authors declare no competing interests.

\paragraph{AI use declaration.}
In this work, we used generative AI tools to implement methods, in the form of writing and editing code. We have not used generative AI tools to generate data sets, formulate mathematical claims or write proofs. Additionally, we used generative AI tools for brainstorming and for drafting and refining the text of this paper. We have reviewed all AI-assisted work: AI-assisted code was reviewed and tested by the authors, and all AI-assisted text was checked and edited by the authors. We take responsibility for the final content of this work, including text, claims or artifacts produced with the aid of generative AI.

\paragraph{Data availability.}
The simulated data used in this study are publicly available as part of the \textsc{ColliderML} dataset~\citep{Elitez2025ColliderML, ColliderMLHF}.

\paragraph{Code availability.}
The code used for training, evaluation and the throughput measurements is available at \url{https://github.com/jonathanrenusch/ssm-colliderml-track-regression}.

\FloatBarrier
\bibliographystyle{unsrtnat}
\bibliography{references}

\appendix
\section{Technical appendices and supplementary material}
\label{app:technical}

\renewcommand{\thefigure}{\Alph{section}.\arabic{figure}}
\renewcommand{\thetable}{\Alph{section}.\arabic{table}}
\renewcommand{\theHfigure}{app.\Alph{section}.\arabic{figure}}
\renewcommand{\theHtable}{app.\Alph{section}.\arabic{table}}
\setcounter{figure}{0}
\setcounter{table}{0}

\subsection{Kernel adaptation for short sequences}
\label{app:kernels}

\cref{tab:kernel-gains} reports single-H100 (NVL) inference throughput
for the three encoders of \cref{fig:encoder-ablation} (trunk-matched at
$\sim\!0.63$\,M parameters) that admit a fused short-sequence kernel,
before and after the treatment of \cref{sec:kernels}, which packs the layout,
fuses the token mixing into one Triton kernel per encoder, compiles the
Fourier encoding and runs the projections in fp16.

\begin{table}[h]
  \centering
  \caption{Single-H100 (NVL) inference throughput, in tracks/s, at the
  saturating batch size ($131\,000$ tracks/batch). The kernel's default
  configuration is: strict IEEE fp32, standard recurrent and SSM kernels,
  full-attention for the Transformer (flash attention does not give any
  gains for short sequences). \emph{Deployed} is the fastest
  physics-gated path: packed batches, a fused Triton kernel for the
  token mixing, a fused Fourier encoding, and fp16 projections for all
  three encoders. All three encoders have the same parameter count
  within 3 \%. \emph{RTX 5000 Ada} reports the same deployed path on a
  single RTX 5000 Ada GPU.}
  \label{tab:kernel-gains}
  \small
  \begin{tabular}{lrrrr}
    \toprule
    encoder & default kernel H100 & optimized H100 & gain H100 & optimized RTX 5000 Ada  \\
    \midrule
    minGRU  & $0.88$\,M & $5.41$\,M & $6.1\times$ & $1.51$\,M \\
    Transformer & $0.48$\,M & $3.18$\,M & $6.6\times$ & $0.855$\,M \\
    Mamba-2  & $0.58$\,M & $2.06$\,M & $3.6\times$ & $0.649$\,M \\
    \bottomrule
  \end{tabular}
\end{table}

\subsection{Learned structure, trunk-gradient cosine probe}
\label{app:interp-grad}

\begin{figure}[!htbp]
  \centering
  \includegraphics[width=0.46\linewidth]{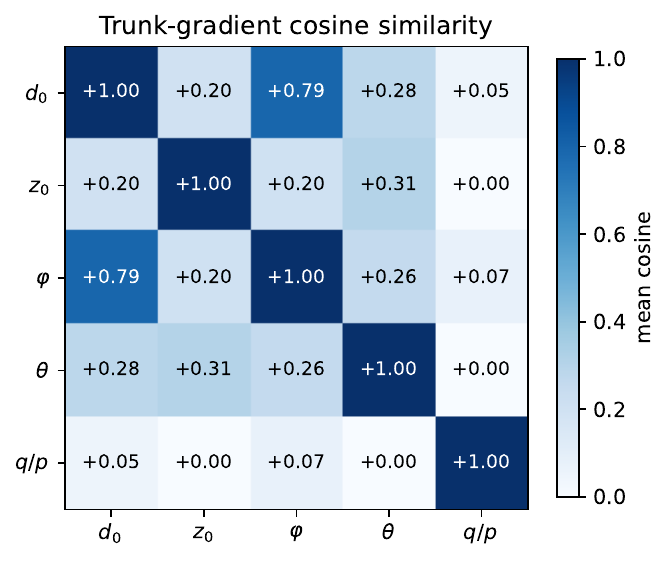}
  \caption{Trunk-gradient cosine similarity of the paper model.
  Mean cosine between the per-parameter loss gradients on the shared
  trunk ($0.63$\,M parameters, output head excluded), averaged over
  $450$ minibatches of $2048$ tracks ($0.92$\,M tracks);
  each cell shows the mean $\pm$ its standard error.  The four
  geometry parameters $(\dz,\zz,\phii,\thetaP)$ are mutually
  aligned, most strongly on the transverse perigee pair
  $(\dz,\phii)$, while $\qop$ is nearly orthogonal to all of
  them.}
  \label{fig:cos_heatmap}
\end{figure}

A trunk-gradient cosine probe (\cref{fig:cos_heatmap}) asks whether
the trained encoder respects the geometric structure of the problem:
for each parameter we backpropagate only that parameter's loss and
measure the alignment of the resulting trunk gradients.  The matrix
separates into a geometry block and a curvature block.  Every pairing among
$(\dz,\zz,\phii,\thetaP)$ is positive and many standard errors away from
zero, led by the transverse perigee pair $(\dz,\phii)$ at
$+0.807\pm0.008$ and the longitudinal lever-arm pair
$(\zz,\thetaP)$ at $+0.315\pm0.011$; the remaining four lie between
$+0.20$ and $+0.25$.  Curvature is nearly orthogonal to all of
them: $|C_{\cdot,\qop}|\le0.048$, with
$C_{\zz,\qop}=+0.002\pm0.003$ and $C_{\thetaP,\qop}=+0.003\pm0.003$
indistinguishable from zero.  The trunk therefore couples exactly
the four parameters that classical perigee geometry couples, and
learns the curvature along a nearly orthogonal direction.
This is physically sensible, since the sagitta is a distinct information
channel from the pointing constraints, and it is consistent with the
scale-free $\qop$ head, whose gradient scale is decoupled by
construction. None of this arises from any supervision on the classical fit, on
the perigee covariance, or on residuals.

\subsection{Encoder comparison in physics performance}
\label{app:arch-ablation}

\begin{figure}[!htbp]
  \centering
  \includegraphics[width=\linewidth]{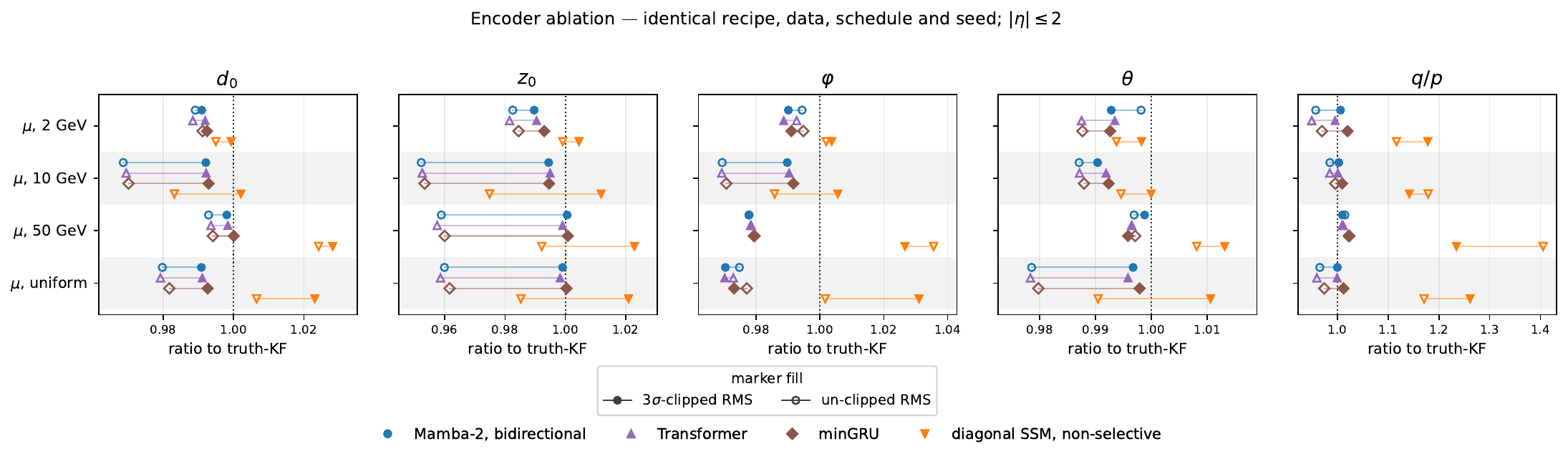}
  \caption{Encoder comparison. Resolution ratio to the truth-seeded \kf{} for
  each perigee parameter (panels) and each muon test sample (rows) at
  $|\eta|\le2$, for four encoders trained with the same recipe, data, schedule,
  random seed and parameter budget ($\sim$$0.63$\,M) and read out through the
  same seed, features and quantile heads. These are first-stage models, trained
  for $25$ epochs at one shared learning rate and without the second-stage training
  of the deployed model. Filled markers are the iterative-$3\sigma$-clipped
  core, open markers the un-clipped \rms{} including all tails. The dotted line
  marks parity with the reference. Three of the four reach the same precision.
  The non-selective diagonal state-space model falls behind, most clearly in
  $\qop$. Single seed per encoder; shared learning rate.}
  \label{fig:encoder-ablation}
\end{figure}

\cref{fig:encoder-ablation} compares the four encoders we tested, all
matched to the minGRU encoder's parameter count to within $3\,\%$ and
trained with the identical recipe, on every muon test sample. The four cover both ways of mixing tokens, recurrence and
attention, and both kinds of state update, input-dependent and fixed: the
Transformer as the general-purpose encoder~\citep{Vaswani2017Attention}, Mamba-2 as the selective state-space model we started our R\&D on \citep{Dao2024Mamba2}, the minGRU as a simpler gated form
\citep{Feng2024minGRU}, and a non-selective diagonal state-space model as a similar recurrence without selectivity. 

\subsection{Results supplement, per-sample figures}
\label{app:results-supplement}

Every resolution figure below is the iterative-$3\sigma$-clipped \rms{}
computed from the tracks estimated by both the minGRU and the truth-seeded
\kf{} shipped with the dataset, one sample per figure, with the
minGRU/\kf{} ratio beneath each panel, inside the acceptance
$|\eta|\le2$ of \cref{sec:results}. Each legend states the unbinned \rms{} and the tracks
removed by the clip; the title states the total tracks; bands are the
analytic \rms{} standard error.  The residual
histograms give, per legend line, the iterative-$3\sigma$ \rms{} and
the clipped fraction.  The fourth
parameter is the polar angle $\theta$; $\eta$ is a binning axis only.

\begin{figure}[!htbp]
  \centering
  \includegraphics[width=\linewidth]{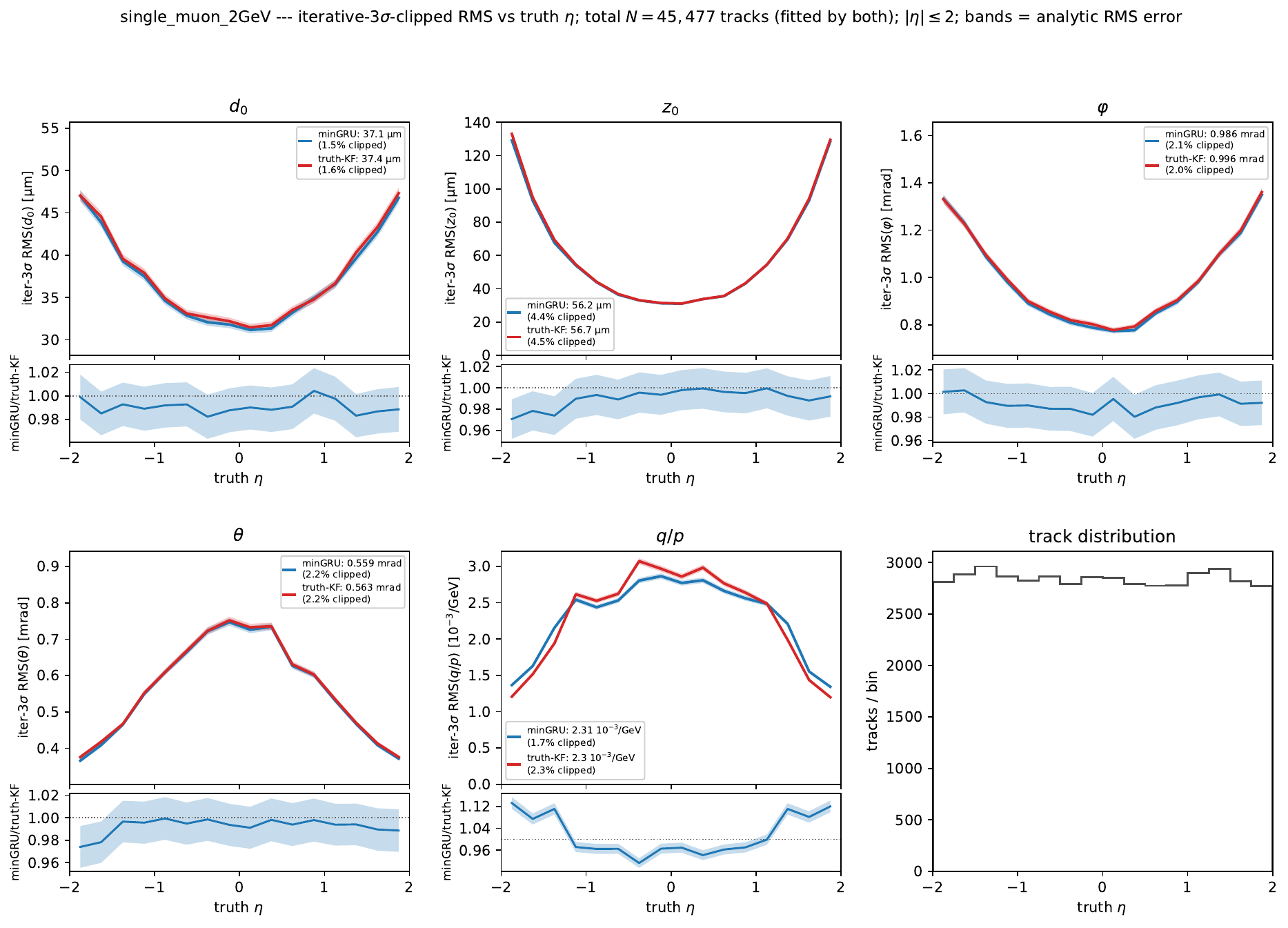}
  \caption{\textbf{single muons, $2$~GeV}, iterative-$3\sigma$-clipped \rms{} versus
  $\eta$ (minGRU/\kf{} ratio beneath each panel; clip fractions in the
  legends, total in the title).}
  \label{fig:app-reta-single_muon_2GeV}
\end{figure}

\begin{figure}[!htbp]
  \centering
  \includegraphics[width=0.72\linewidth]{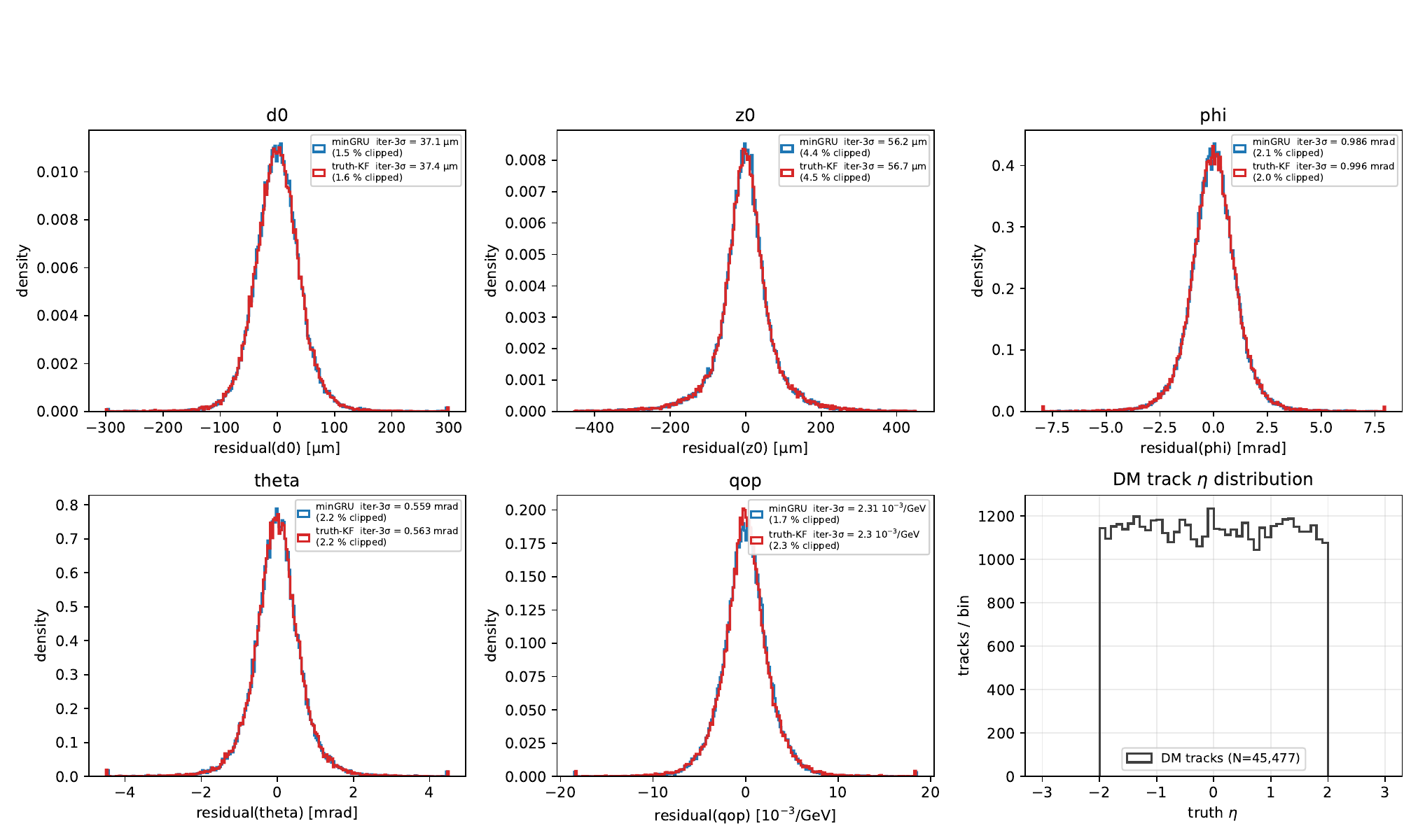}
  \caption{\textbf{single muons, $2$~GeV}, residual distributions; each legend gives the
  iterative-$3\sigma$ \rms{} and the clipped fraction.}
  \label{fig:app-resid-single_muon_2GeV}
\end{figure}

\begin{figure}[!htbp]
  \centering
  \includegraphics[width=0.72\linewidth]{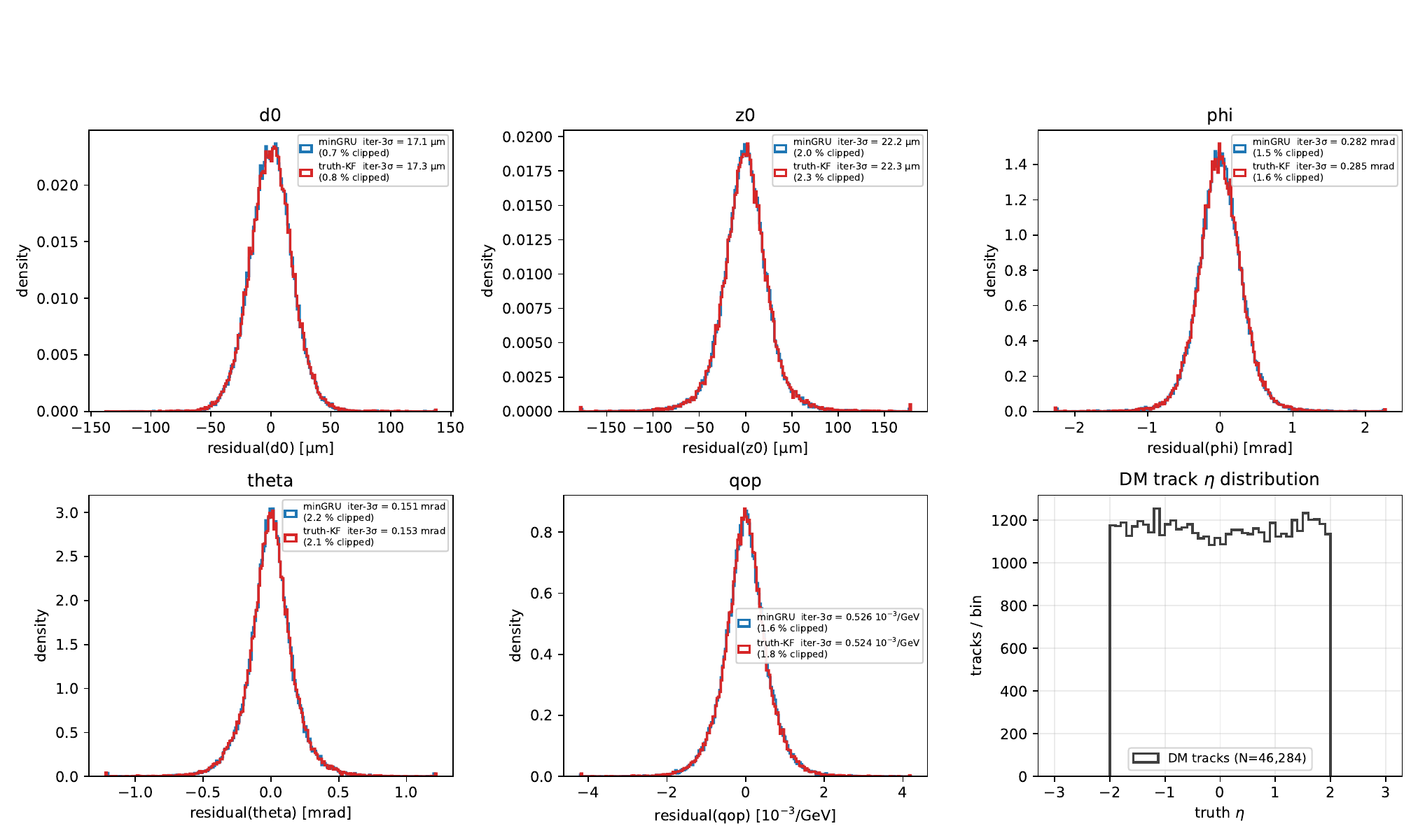}
  \caption{\textbf{single muons, $10$~GeV}, residual distributions; each legend gives the
  iterative-$3\sigma$ \rms{} and the clipped fraction.  (The
  \SI{10}{\GeV} resolution-versus-$\eta$ curves are the main-text
  \cref{fig:rms_eta}.)}
  \label{fig:app-resid-single_muon_10GeV}
\end{figure}

\begin{figure}[!htbp]
  \centering
  \includegraphics[width=\linewidth]{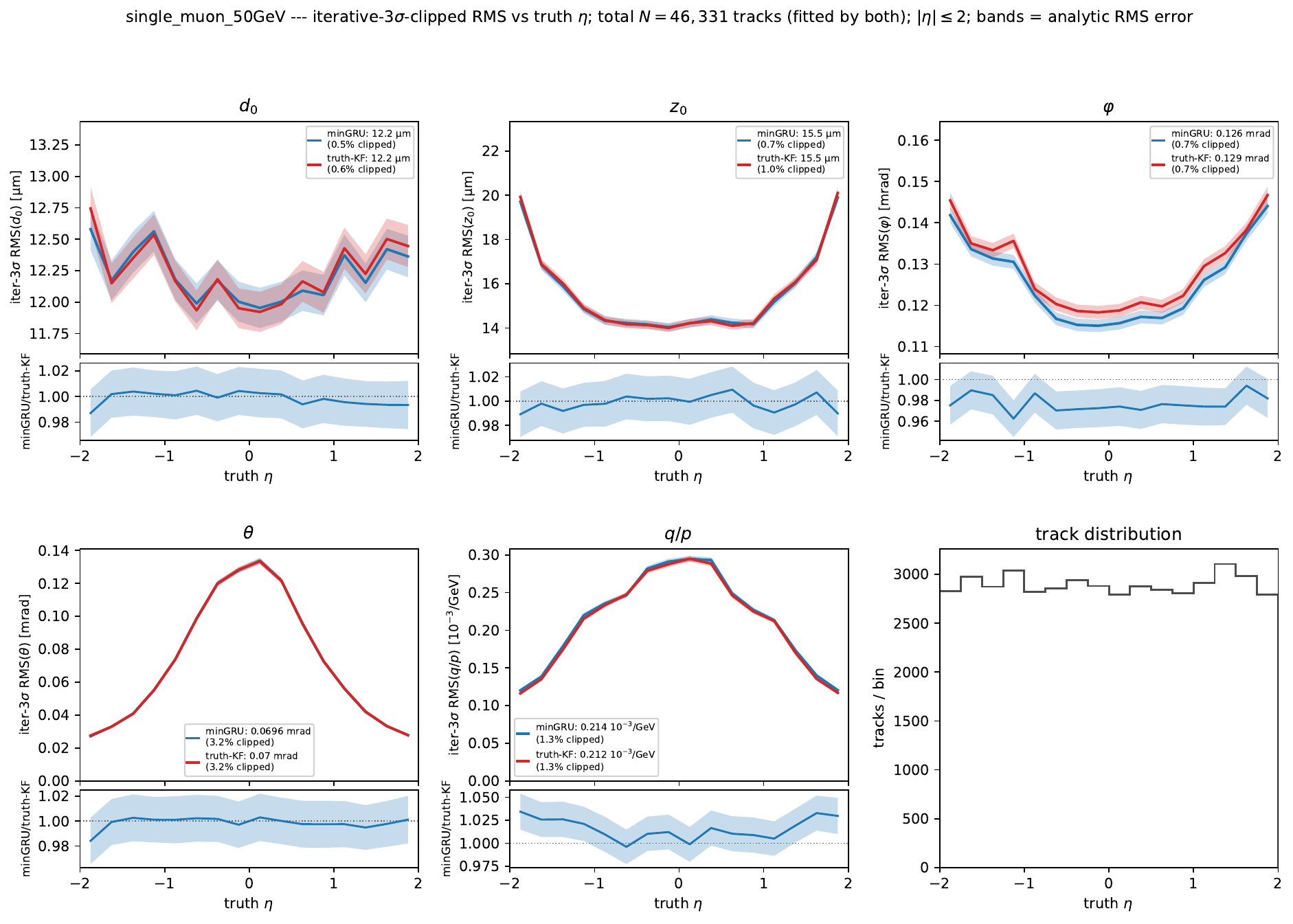}
  \caption{\textbf{single muons, $50$~GeV}, iterative-$3\sigma$-clipped \rms{} versus
  $\eta$ (minGRU/\kf{} ratio beneath each panel; clip fractions in the
  legends, total in the title).}
  \label{fig:app-reta-single_muon_50GeV}
\end{figure}

\begin{figure}[!htbp]
  \centering
  \includegraphics[width=0.72\linewidth]{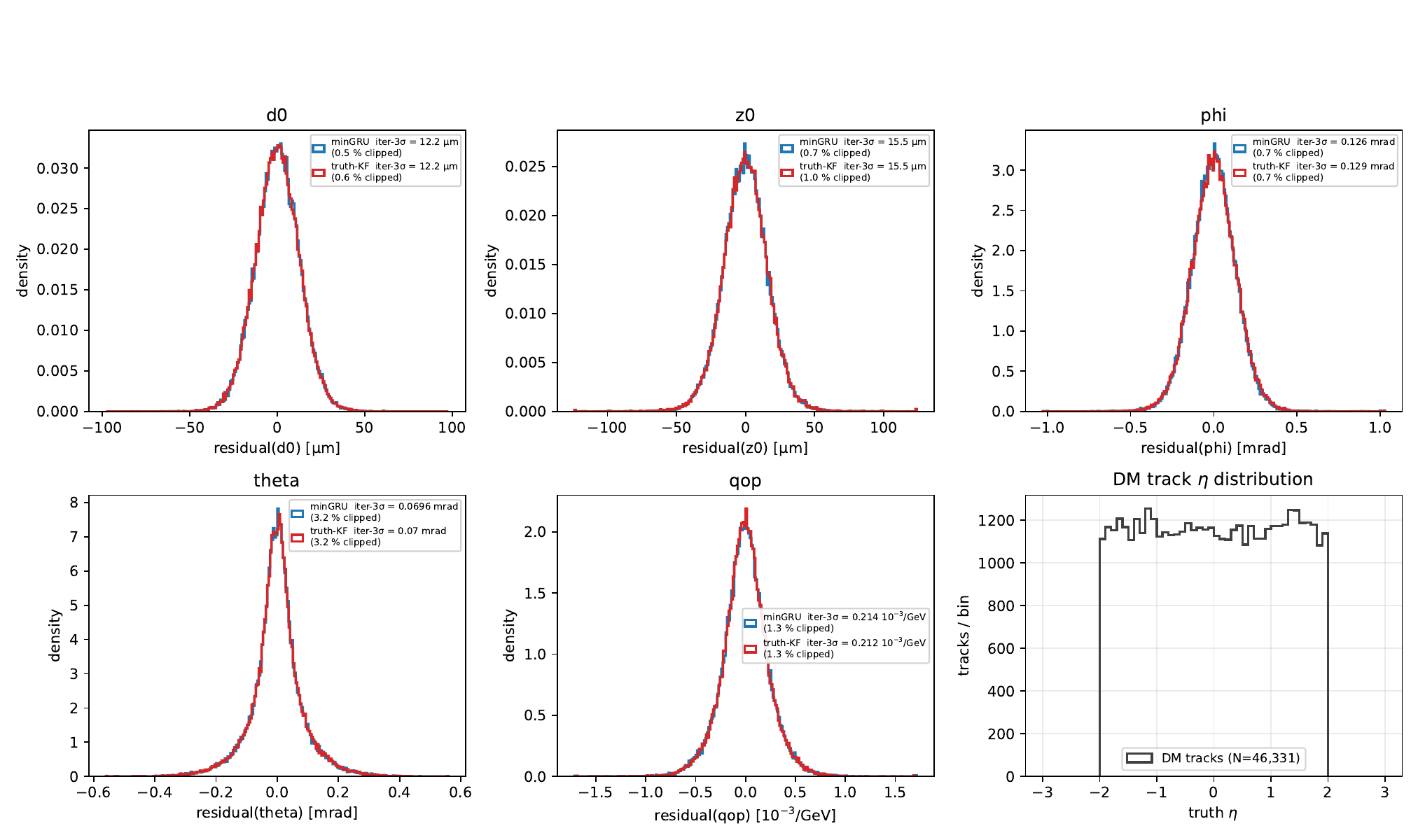}
  \caption{\textbf{single muons, $50$~GeV}, residual distributions; each legend gives the
  iterative-$3\sigma$ \rms{} and the clipped fraction.}
  \label{fig:app-resid-single_muon_50GeV}
\end{figure}

\begin{figure}[!htbp]
  \centering
  \includegraphics[width=\linewidth]{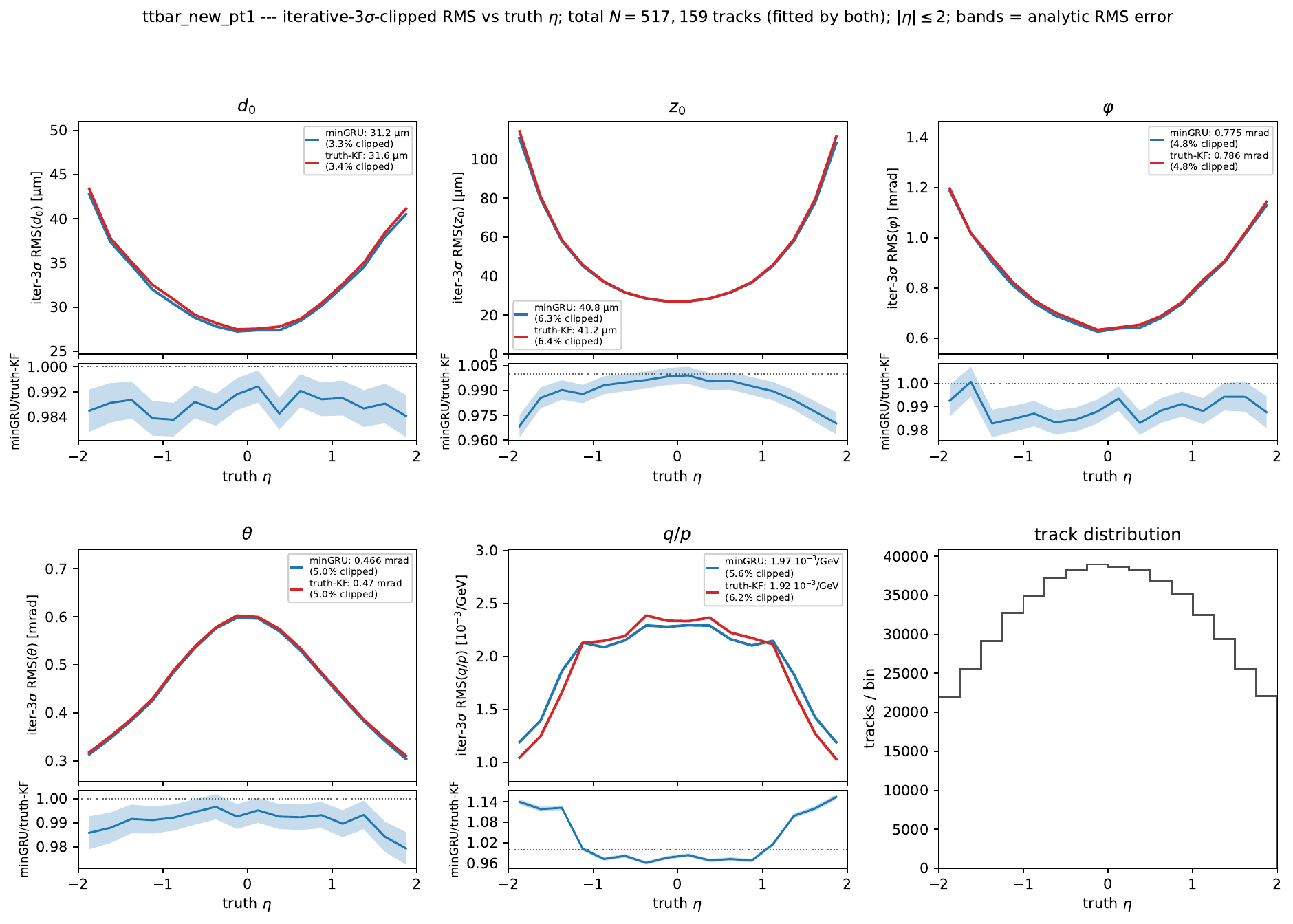}
  \caption{\textbf{\ttbar{} events, $p_{\mathrm T}>1$~GeV}, iterative-$3\sigma$-clipped \rms{} versus
  $\eta$ (minGRU/\kf{} ratio beneath each panel; clip fractions in the
  legends, total in the title).}
  \label{fig:app-ttbar}
\end{figure}

\end{document}